\documentclass{article}

\usepackage{PRIMEarxiv}

\usepackage[utf8]{inputenc}
\usepackage[T1]{fontenc}

\usepackage{hyperref}
\usepackage{url}
\usepackage{graphicx}
\usepackage{booktabs}
\usepackage{amsfonts}
\usepackage{nicefrac}
\usepackage{microtype}
\usepackage{fancyhdr}
\usepackage{multirow}
\usepackage{makecell}
\usepackage{amsmath,amssymb,mathtools}
\usepackage{algorithm}
\usepackage{algpseudocode}
\usepackage{wrapfig}
\usepackage{subcaption}
\usepackage{xspace}

\newcommand{\ie}{i.e.\xspace}
\newcommand{\etal}{et al.\xspace}

\newcommand{\RR}{\mathrm{RR}}
\newcommand{\PARAM}{\textsc{PARAM}\xspace}
\newcommand{\WIP}{\textsc{WIP}\xspace}

\begin{document}

\title{Improving Relative Representations with Learned Anchors and Whitened Inner Products}

\author{
Oscar Thorsted Svendsen$^{*,1}$ \quad
Nikolaj Holst Jakobsen$^{*,1}$ \quad
Fabian Mager$^{\dagger,1}$ \quad
Hiba Nassar$^{\dagger,1}$\\
$^1$Technical University of Denmark\\
\texttt{\{s224177,s234818,hibna,fmager\}@dtu.dk}\\[0.5em]
\small $^*$These authors contributed equally to this work and share first authorship.\\
\small $^\dagger$These authors advised the project.
}

\maketitle

\begin{abstract}
Independently trained neural models typically converge to incompatible latent representations, creating a fundamental barrier to highly modular AI systems. While Relative Representations (RR) address this by mapping absolute coordinates to a shared space defined by similarities to common anchor points, traditional implementations rely on randomly sampled anchors and cosine similarity, which frequently fail to capture the anisotropic geometries of modern architectures like Transformers. In this work, we propose a robust framework for cross-model communication based on two improvements. We learn anchors as robust semantic prototypes and utilize a geometry-aware similarity metric which preserves discriminative magnitude information and is invariant to affine shifts. Our approach demonstrates significant gains in performance and consistency across vision and language tasks. Notably, it enables nearly lossless information transfer and stable zero-shot communication even between highly heterogeneous architectures, such as small language models of varying scales.

\end{abstract}
\section{Introduction}
\label{sec:intro}

Relative Representations (RR) address a central problem in modular representation learning: independently trained encoders can capture the same underlying semantics while organizing their latent spaces differently. Even when two models represent the same inputs, their embedding spaces may differ by rotation, anisotropic scaling, or translation, making direct transfer between them difficult. RR avoids relying on absolute coordinates by representing each embedding through its similarities to a shared set of \emph{anchors}. For example, the word \textit{``king''} should be similar to \textit{``queen''} and \textit{``castle''} across English, French, and German, even if the absolute embeddings differ. RR extends this idea to general embedding spaces by using anchor-based similarities as a common coordinate system.

We formalize this setting as follows: we consider $\mathcal{E} = \{\mathbf{E_1}, \dots, \mathbf{E_B}\}$ to be a set of $B$ independently trained encoders. For any encoder $E_i \in \mathcal{E}$, let $\mathbf{X}^i = \{x^i_n\}_{n=1}^N \subset \mathbb{R}^d$ denote the set of $d$-dimensional embeddings\footnote{For readability, we write the embedding dimension as $d$ throughout; in general it may depend on the encoder ($d=d_i$), and all definitions extend by replacing $d$ with $d_i$ where appropriate.} produced for N parallel inputs, meaning the same underlying input embedded in each space. We define an anchor set $\mathbf{A} = \{a_1,\cdots, a_m\}$. Given a similarity function $s(\cdot, \cdot)$, the RR mapping for an embedding $x \in \mathbb{R}^d$ is:
\begin{equation}
    RR(x) = [s(x, a_1), \dots, s(x, a_m)]^\top \in \mathbb{R}^m    
\end{equation}
where $s(\cdot,\cdot)$ takes two vectors as input and returns a single scalar. In the standard formulation proposed by Moschella \etal\cite{moschella2023relative}, anchors are a randomly sampled subset of parallel datapoints $\mathbf{A} \subset \mathbf{X}$ and cosine similarity is used as $s(\cdot,\cdot)$. If two independently trained encoders use the same anchors and similarity, their RR features can be sufficiently aligned to enable \emph{zero-shot stitching}, \ie, training a module on relative embeddings from one encoder and deploying it on relative embeddings from another.

The goal of RR is to provide a shared relative interface across independently trained encoders. For this to hold, the relative mapping should satisfy the following two objectives:

\paragraph{Cross-space alignment} ($O_{al}$): For a set of $N$ parallel inputs, the relative coordinates produced by any two encoders $E_i$ and $E_j$ must align such that the distance over the sum of paired points is minimized: $\sum_{n=1}^{N} d(RR(x_n^i), RR(x_n^j)) \approx 0$, where $d(\cdot, \cdot)$ is a distance metric.

\paragraph{Information Preservation} ($O_{inf}$): The relative features should retain the task-relevant signal present in the absolute embeddings, so that a downstream model trained on $RR(x)$ can achieve performance comparable to one trained directly on $\mathbf{X}$.

\begin{figure}
    \centering
    \includegraphics[width=1\linewidth]{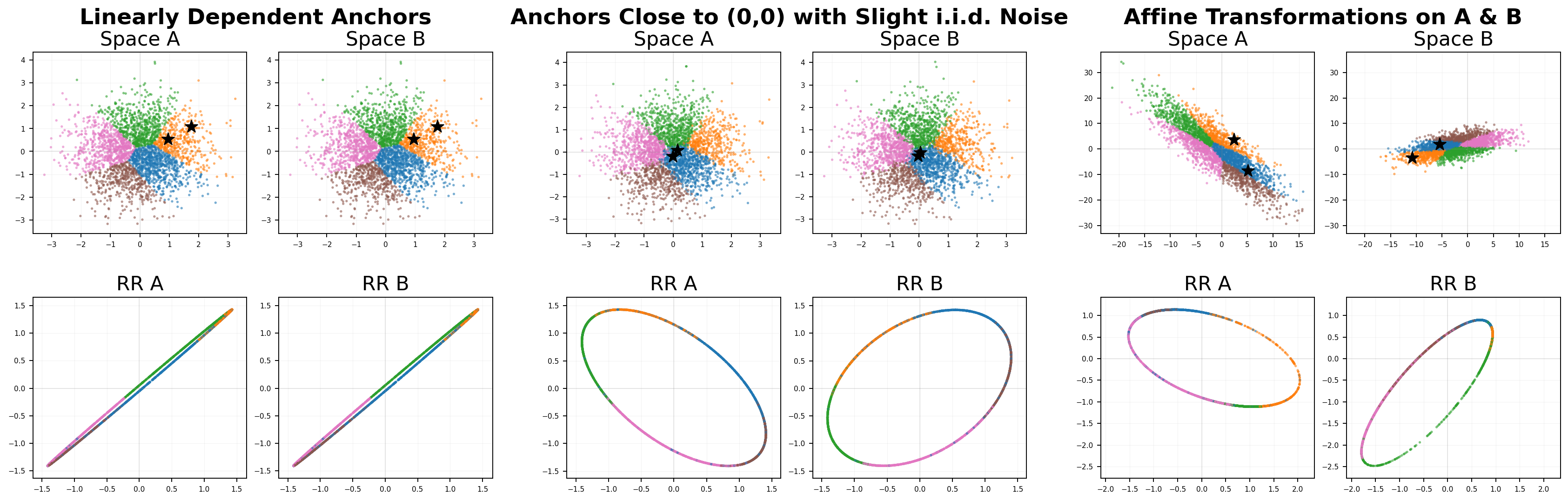}
    \caption{Shows conditions that may arise using cosine similarity as $s(\cdot,\cdot)$ and random anchors and their effect on RR: i) The dimensionality collapse of linearly dependent anchors. ii) How $s(\cdot,\cdot)$ amplifies noise in anchors close to $(0,0)$. iii) $s(\cdot,\cdot)$ invariance to certain affine transformations.}
    \label{fig:rr_problem_illustration}
\end{figure}

Standard RR can fall short on both objectives for two reasons. First, anchors are usually chosen by random sampling, which provides no guarantee of good coverage of the data manifold. Poorly distributed or redundant anchors can yield low-rank relative features and discard task-relevant information, compromising $O_{inf}$. Second, cosine similarity depends only on angular relationships, discarding magnitude information that can carry useful signal such as feature confidence \cite{pmlr-v267-draganov25a}. It can also be unstable under mean shifts, and anisotropic scaling which is common in transformer-based embedding spaces \cite{ethayarajh2019contextual,godey2024anisotropy}. Moreover, near the origin, cosine similarity becomes noise-sensitive: small perturbations in a vector can cause disproportionately large changes in its normalized direction. As a result, even with shared anchors, relative coordinates may shift across encoders, weakening $O_{al}$ and reducing $O_{inf}$, illustrated in Fig.~\ref{fig:rr_problem_illustration}.

These limitations motivate two design hypotheses. First, to support both $O_{al}$ and $O_{inf}$, anchors should be informative and stable: they should cover the manifold, avoid redundancy, and preserve a consistent semantic role across embedding spaces. Second, to support $O_{al}$, the similarity function should be robust to the affine distortions that commonly separate independently trained latent spaces, while still preserving useful angular and magnitude information. The method proposed in this paper addresses these two requirements jointly through learned anchor construction (\PARAM) and a covariance-aware similarity measure (\WIP).

\section{Related Work}
\label{sec:related}

Moschella \etal \cite{moschella2023relative} introduced Relative Representations (RR) for zero-shot stitching through shared anchors, a framework recently extended for latent translation by Maiorca \etal\cite{maiorca2024latent}. While these works explored anchor selection methods like k-means and farthest-point sampling (FPS), these strategies are fundamentally limited to static, discrete points and leave room for improvement elsewhere. Our work builds directly on this framework: we keep the same overall idea of representing each datapoint by its similarities to anchors, but revisit the two design choices that most directly determine the quality of the resulting relative space, namely anchor construction and similarity.

A basic reason RR is plausible is that independently trained models often learn representations with related geometric structure. Kornblith \etal~\cite{kornblith2019similarity} showed that representations learned by different neural networks can still exhibit reliable similarities, even across different random initializations. In our setting, this motivates creating a shared relative space system rather than relying on absolute latent coordinates -- taking inspiration from Huh \etal \cite{huh2024platonic}.

Among follow-up works, the most relevant to our study is H{\"u}ttebr{\"a}ucker \etal~\cite{huttebraucker2024semantic_equalization}, who apply RR to semantic channel equalization and argue that anchor selection is important for performance. Their work is especially relevant because it moves beyond purely random anchors and proposes a more structured anchor-selection strategy. This aligns closely with our motivation: if the relative-space dimensionality equals the number of anchors, then anchor redundancy or poor coverage directly limits the quality of the representation. Our contribution differs in that we do not select anchors as fixed points; instead, \PARAM learns them as convex mixtures of parallel datapoints.


\section{Method}
\label{sec:method}
To address the limitations of standard RR and satisfy the objectives of cross-space alignment ($O_{al}$) and information preservation ($O_{inf}$), we propose a framework consisting of two components: (i) \emph{Parameterized Anchor Representations via Adaptive Matrices (PARAM)}, which learns anchors as convex mixtures of parallel datapoints, (ii) and the \emph{Whitened Inner Product (WIP)}, a covariance-aware similarity designed to be robust to affine distortions, to make relative coordinates more stable across embedding spaces.

\subsection{PARAM: Learning Anchors as Convex Mixtures of Datapoints}

To ensure the relative coordinate system is built upon a robust foundation we propose PARAM to move beyond the limitations of treating anchors as a fixed subset of the data manifold. Instead of sampling a static set $\mathbf{A} \subset \mathbf{X}$, PARAM defines anchors as learnable convex mixtures of a support set, optimizing their placement to maximize information preservation and cross-space consistency.

Let $\mathbf{X}_{\text{sub}}^{i} \in \mathbb{R}^{K \times d}$ denote a "support set" comprising a subset of $K$ parallel embeddings $\mathbf{X}_{\text{sub}}^i \subseteq \mathbf{X}^i$ from each encoder $E_i \in \mathcal{E}$. We define the set of $m$ anchors for the $i$-th embedding space as a matrix $\mathbf{A}^i = [a_1^i, \dots, a_m^i]^\top \in \mathbb{R}^{m \times d}$. Each individual anchor $a_r^i \in \mathbb{R}^d$ (the $r$-th row of $\mathbf{A}^i$) is generated through a shared mixture matrix $\mathbf{P}$:

\begin{equation}
\mathbf{A}^{i} = \mathbf{P}\mathbf{X}^{i}_{\text{sub}} \qquad \mathbf{P}\in\mathbb{R}^{m\times K}, 
\label{eq:PARAM}
\end{equation}
where $\mathbf{P}$ is a shared weight matrix optimized across all embedding spaces. To ensure anchors remain semantically grounded within the convex hull of the data, $\mathbf{P}$ is constrained to be row-stochastic. Applying $\mathbf{P}$ to the corresponding support sets $\mathbf{X}_{sub}^i$ of all $B$ encoders creates a semantically aligned anchor set across spaces: $\mathcal{A} = \{\mathbf{A}^1, \dots, \mathbf{A}^B\}$.

Constructing anchors as convex mixtures provides a significant denoising effect through averaging. If we consider encoder-specific differences in embedding positions as approximately i.i.d. noise across individual points, the averaging operation $a_{r}^{i} = \sum_{j=1}^K P_{rj} (\mathbf{X}_{\text{sub}}^i)_j$ reduces the variance of the resulting anchors. Specifically, if the learned weights $\mathbf{P}$ are uniform, this provides a $1/\sqrt{K}$ reduction in noise standard deviation. This ensures that each anchor $a_r^i$ acts as a stable, weighted prototype that is less sensitive to the stochastic variations or architectural biases of any single $E_i$. Furthermore, PARAM enables continuous anchor placement within the latent space, unlike traditional methods that restrict anchors to discrete, existing datapoint.

\subsection{Training Objectives}
\label{sec:training_objectives}

We optimize the mixture matrix $\mathbf{P}$ by minimizing a weighted sum of losses. We separate objectives that can be computed within a single embedding space from objectives that require paired samples across embedding spaces. Full definitions and equations are given in Appendix~\ref{app:losses}.

\paragraph{Single-space objectives:}
\begin{itemize}
    \item \emph{Coverage} $(\mathcal{L}_{\mathrm{cov}})$: soft $k$-means clustering which pulls anchors toward centroids to cover the space, reducing risk of redundancy/collapse.
    \item \emph{Orthogonality} $(\mathcal{L}_{\mathrm{orth}})$: penalizes correlated anchors, improving effective rank of RR features and reducing degenerate coordinates.
    \item \emph{Length control} $(\mathcal{L}_{\mathrm{len}})$: constrains anchors to have controlled $L_2$ norm, preventing collapse toward the origin or drifting to extreme scales.
\end{itemize}

\vspace{0.3cm}
\paragraph{Multi-space objective:}
\begin{itemize}
    \item \emph{Symmetric InfoNCE (Information Noise-Contrastive Estimation)} $(\mathcal{L}_{\mathrm{symNCE}})$: cross-space alignment of RR features for parallel points across embedding spaces, pulling corresponding samples together and pushing other samples apart \cite{oord2018cpc}.
\end{itemize}

\paragraph{Total objective}
With weights $\{\lambda_\cdot\}$, we minimize
\begin{equation}
    \mathcal{L}=\lambda_{\mathrm{symNCE}}\mathcal{L}_{\mathrm{symNCE}}
+\lambda_{\mathrm{cov}}\mathcal{L}_{\mathrm{cov}}
+\lambda_{\mathrm{len}}\mathcal{L}_{\mathrm{len}}
+\lambda_{\mathrm{orth}}\mathcal{L}_{\mathrm{orth}}.
\end{equation}

If $B=1$, we train \PARAM using only $\{\mathcal{L}_{\mathrm{cov}},\mathcal{L}_{\mathrm{len}},\mathcal{L}_{\mathrm{orth}}\}$.
If $B>1$, we additionally include $\mathcal{L}_{\mathrm{symNCE}}$ to align RR spaces across encoders using paired samples.

\subsection{Similarity: Whitened Inner Product (\WIP)}
\label{sec:wip}

To satisfy Cross-Space Alignment ($O_{al}$) and Information Preservation ($O_{inf}$), the similarity function $s(\cdot,\cdot)$ must be robust to the global mean shifts and anisotropic scaling that characterize independently trained encoders. We propose the Whitened Inner Product (WIP), which represents the similarity between an embedding $x$ and an anchor $a_r^i$ as a standard inner product within a whitened coordinate system.

Formally, for each embedding space $i$, we first remove the empirical mean $\mu^i$ and apply a symmetric whitening transformation $\mathbf{L}^i = (\boldsymbol{\Sigma}^i)^{-1/2}$. The WIP similarity is then defined as:
\begin{equation}
    s_{\text{WIP}}(x, a_r^i) = \langle \mathbf{L}^i(x - \mu^i), \mathbf{L}^i(a_r^i - \mu^i) \rangle = (x - \mu^i)^\top (\boldsymbol{\Sigma}^i)^{-1} (a_r^i - \mu^i)
\end{equation}
\label{eq:wip}
This directly works in favor of $O_{al}$ by making the similarity less sensitive to encoder-specific affine distortions. Exact affine invariance holds only for the non-shrunk form (Appendix~\ref{app:wip}). In addition, WIP preserves both the magnitude and angular information of the embeddings, which is critical for $O_{inf}$ when norms carry discriminative signal.

\begin{figure}[h!]
  \centering
  \includegraphics[width=\linewidth]{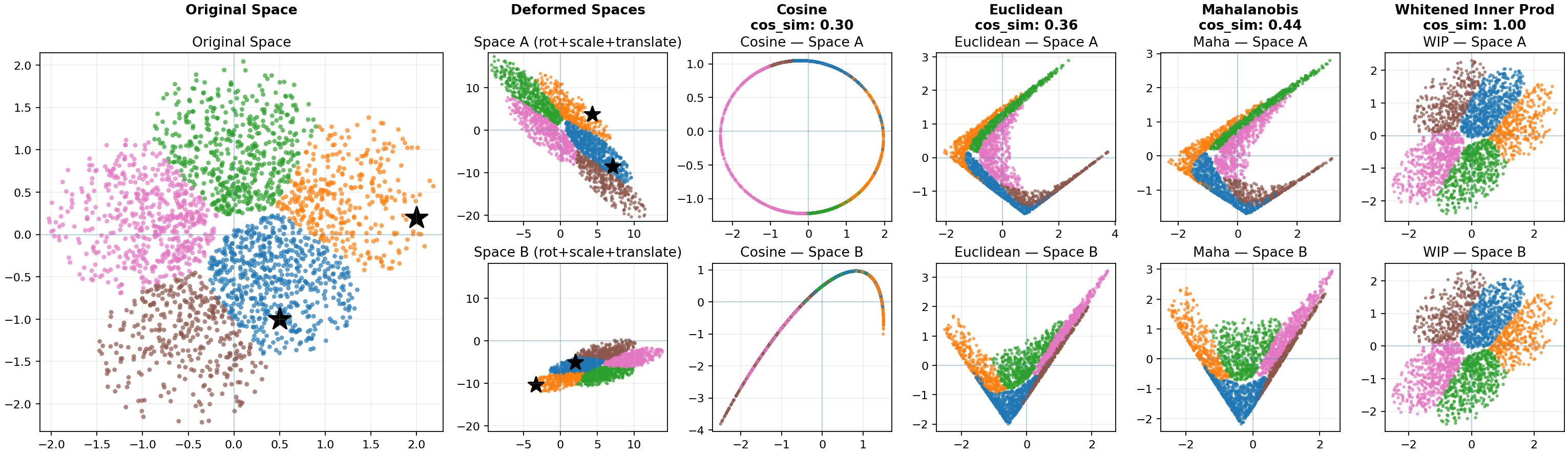}
  \caption{
  Effect of $s(\cdot,\cdot)$ on $\RR$ geometry under cross-space distortions.
  Left: original clustered space and two deformed versions (rotation + scaling + translation).
  Right: $\RR$ coordinates computed with different similarities and the cosine similarity between the resulting $\RR$ spaces.
  }
  \label{fig:wip_geometry}
\end{figure}

Figure \ref{fig:wip_geometry} demonstrates the impact of $s(\cdot,\cdot)$ on RR geometry. Euclidean distance based measures tend to create distorted spaces heavily affected by the distance between the anchors. Cosine similarity results in a warped space where all datapoints lie on an approximately elliptical shell (when $d\le m$). In contrast, WIP  yields a more consistent cluster geometry across the deformed embedding spaces.

\section{Implementation and Zero-Shot Pipeline}
\label{sec:Implementation}

We evaluate \emph{zero-shot stitching} by holding out a test embedding space during optimization of the mixture matrix $\mathbf{P}$, then deploying a classifier trained on relative features from other embedding spaces directly on the held-out embedding space without fine-tuning. The protocol can be seen in Algorithm \ref{alg:zs_pipeline}

\begin{algorithm}[h]
\caption{Training \PARAM and zero-shot stitching with \WIP}
\label{alg:zs_pipeline}
\begin{algorithmic}[1]
\Require Embedding spaces $\{\mathbf{X}^{1},\cdots, \mathbf{X}^B\}$ for the same $N$ inputs; support subsets $\{\mathbf{X}^{1}_{\text{sub}},\cdots, \mathbf{X}^B_{\text{sub}}\}$; number of anchors $m$; similarity $s_{\WIP}$ as defined in Eq.~\ref{eq:wip}
\Ensure Learned mixture matrix $\mathbf{P}$; zero-shot predictions on held-out test space $\mathbf{X}^{t}$.
\State Optimize $\mathbf{P}$ across all embedding spaces, excluding $\mathcal{L}_{\mathrm{symNCE}}$ if $B=1$.

\For{each $i \in \{1,\dots,B\}\setminus\{t\}$}
    \State Construct anchors $\mathbf{A}^{i} \leftarrow \mathbf{P}\,\mathbf{X}^{i}_{\text{sub}}$.
    \State Compute $\mathbf{X}^{i}_{\RR} \leftarrow \{\RR(x)\}_{x\in \mathbf{X}^{i}}$
    \State \hspace{1.6em}where $\RR(x)=\big[s_{\WIP}(x,a^{i}_{1}),\dots,s_{\WIP}(x,a^{i}_{m})\big]_{a\in \mathbf{A}^{i}}$.
\EndFor

\State Construct anchors $\mathbf{A}^{t} \leftarrow \mathbf{P}\,\mathbf{X}^{t}_{\text{sub}}$.
\State Compute $\mathbf{X}^{t}_{\RR} \leftarrow \{\RR(x)\}_{x\in \mathbf{X}^{t}}$
\State \hspace{1.6em}where $\RR(x)=\big[s_{\WIP}(x,a^{t}_{1}),\dots,s_{\WIP}(x,a^{t}_{m})\big]_{a\in \mathbf{A}^{t}}$.

\For{each $i \in \{1,\dots,B\}\setminus\{t\}$}
    \State Train a classifier $h^{i}$ on $(\mathbf{X}^{i}_{\RR},\,y)$.
    \State Evaluate $h^{i}$ on the held-out test features $\mathbf{X}^{t}_{\RR}$ (zero-shot).
\EndFor
\end{algorithmic}
\end{algorithm}
\newcommand{\enc}[1]{\texttt{#1}}
\section{Experiments}
\label{sec:experiments}

Our experimental framework evaluates zero-shot latent space communication through a "stitched" model architecture. Following the pipeline defined in Algorithm \ref{alg:zs_pipeline}, encoder that embeds data points and a \emph{relative} decoder trained on the RR embeddings from a different encoder. Crucially, to ensure a true zero-shot setting as in \cite{moschella2023relative}, PARAM is never trained on the test-encoder, and the stitching operation is always performed without fine-tuning or adapting the decoder. In all experiments random anchors with cosine similarity as described in \cite{moschella2023relative} serves as a baseline.

\subsection{Image and Text Classification}
Like in \cite{moschella2023relative} we evaluate our framework on two standard classification benchmarks to test its robustness across different modalities and architectures. 

\begin{itemize}
    \item \textbf{Image Classification:} We perform a zero-shot classification task on on CIFAR-100 (coarse; 20 labels) \cite{krizhevsky2009cifar}, using 4 different pretrained encoders: Three variants of the ViT transformer \cite{dosovitskiy2021vit} and one variant of the RexNet CNN\cite{han2020rexnet}. This selection specifically probes the framework's ability to align near-isotropic CNN spaces with the highly anisotropic, "cone-shaped" distributions characteristic of latent spaces from transformers \cite{ethayarajh2019contextual,godey2024anisotropy}.
    \item \textbf{Text Classification:} We perform a zero-shot cross-lingual stitching task on the coarse Amazon Reviews Multilingual dataset (MARC) \cite{keung2020marc}, a 2-class product-review sentiment benchmark.  Each language is represented by its own pretrained BERT-variant model \cite{devlin2019bert}: \texttt{bert-base-uncased} for English, \texttt{bert-base-german-cased} for German, \texttt{camembert-base} for French, \texttt{bert-\\base-spanish-wwm-cased} for Spanish, and \texttt{bert-base-chinese} for Chinese.
\end{itemize} 

In both settings, we compare the baseline against two proposed configurations: PARAM Single, where $\mathbf{P}$ is optimized using single-space losses $\{\mathcal{L}_{cov}, \mathcal{L}_{len}, \mathcal{L}_{orth}\}$, and PARAM Multi, which incorporates $\mathcal{L}_{symNCE}$ across all encoders, $\mathcal{E}$, except the test encoder.

\subsection{Zero-Shotting Language Models}
To test the robustness of our framework, we perform zero-shot stitching between architecturally different SLMs (\enc{GPT-2, TinyLlama, Gemma}) ranging from 124M to 1.5B parameters. The dataset consists of $3300$ template-generated English sentences. Using last-token hidden states from synthetically generated sentences, we compute RR for each model $E_i$ and evaluate cross-model alignment via nearest-neighbor retrieval. For every ordered pair of models $(E_i, E_j)$, we measure whether each test sentence's $RR$ produced by $E_i$ retrieves the \emph{same sentence's} $RR$ produced by $E_j$ as its nearest neighbor measured by cosine similarity ($R@1$ \& $R@5$). Unlike the classification experiments, this evaluation requires no decoder training, and directly quantifies the geometric alignment of the $RR$ spaces across architectures with vastly different hidden dimensionalities $d_i \in [768, 2048]$.

\section{Results}
\label{sec:results}
Table \ref{tab:CIFAR_classification_results} shows that both PARAM variants significantly outperform the Random-Cosine baseline across all encoder pairings. This gap is most pronounced when stitching the CNN-based RexNet embeddings to the Transformer-based ViT decoders. PARAM Single outperforms with matching encoder and decoder and achieves a near-equal F1 score with absolute classification. Additionally, the zero-shot performance for each encoder is \emph{also} consistently near the absolute, indicating that PARAM enables nearly lossless transfer of task-relevant information across latent spaces.

\begin{table}[h!]
\caption{Stitching performance with CNN- and Transformer-based encoders on CIFAR-100 (coarse).
Mean-weighted F1 ($\pm$ std) across 6 seeds with $m{=}300$ anchors.
Rows indicate the \emph{test encoder} (where evaluation happens). The \emph{classifier} column indicates the encoder whose RR features the classifier was trained on. The classifiers are \enc{RexNet-100}, \enc{ViT-small-patch16-224}, \enc{ViT-base-patch16-224}, \enc{ViT-base-resnet50-384}.}
\centering
\footnotesize
\setlength{\tabcolsep}{5pt}
\begin{tabular}{ll c ccc}
\toprule
& & \multicolumn{1}{c}{Absolute} & \multicolumn{3}{c}{Relative} \\
\cmidrule(lr){3-3}\cmidrule(lr){4-6}
Encoder & Classifier &
Abs\textbf{.} &
\makecell{Random-Cos} &
\multicolumn{2}{c}{PARAM-WIP} \\
\cmidrule(lr){5-6}
& & & & Multi & Single \\
\midrule

\multirow{4}{*}{\enc{ViT-S}} 
& \enc{ViT-S}  & $90.26 \pm 0.26$  & $87.92 \pm 0.19$ & $89.59 \pm 0.19$ & $90.00 \pm 0.09$ \\
& \enc{RexNet} & --                & $71.11 \pm 1.28$ & $88.05 \pm 0.17$ & $86.86 \pm 0.28$ \\
& \enc{ViT-B}  & --                & $75.30 \pm 1.52$ & $88.03 \pm 0.24$ & $87.27 \pm 0.21$ \\
& \enc{ViT-R50} & --               & $75.48 \pm 1.64$ & $88.14 \pm 0.18$ & $87.60 \pm 0.20$ \\
\midrule

\multirow{4}{*}{\enc{RexNet}} 
& \enc{ViT-S}  & --                & $56.67 \pm 2.11$ & $78.64 \pm 0.26$ & $76.94 \pm 0.22$ \\
& \enc{RexNet} & $81.64 \pm 0.13$  & $76.79 \pm 0.22$ & $80.94 \pm 0.19$ & $81.03 \pm 0.16$ \\
& \enc{ViT-B}  & --                & $53.07 \pm 1.48$ & $78.25 \pm 0.30$ & $76.61 \pm 0.24$ \\
& \enc{ViT-R50} & --               & $53.41 \pm 1.78$ & $78.35 \pm 0.09$ & $77.19 \pm 0.17$ \\
\midrule

\multirow{4}{*}{\enc{ViT-B}} 
& \enc{ViT-S}  & --                & $82.28 \pm 0.80$ & $91.94 \pm 0.10$ & $91.76 \pm 0.13$ \\
& \enc{RexNet} & --                & $77.70 \pm 0.96$ & $91.63 \pm 0.16$ & $90.75 \pm 0.23$ \\
& \enc{ViT-B}  & $93.24 \pm 0.14$  & $91.50 \pm 0.18$ & $92.90 \pm 0.11$ & $93.15 \pm 0.10$ \\
& \enc{ViT-R50} & --               & $80.84 \pm 1.12$ & $91.95 \pm 0.14$ & $91.77 \pm 0.05$ \\
\midrule

\multirow{4}{*}{\enc{ViT-R50}} 
& \enc{ViT-S}  & --                & $82.25 \pm 0.51$ & $89.73 \pm 0.20$ & $89.77 \pm 0.10$ \\
& \enc{RexNet} & --                & $76.67 \pm 1.01$ & $89.40 \pm 0.06$ & $88.96 \pm 0.16$ \\
& \enc{ViT-B}  & --                & $80.70 \pm 0.90$ & $89.72 \pm 0.08$ & $89.68 \pm 0.15$ \\
& \enc{ViT-R50} & $91.48 \pm 0.10$ & $89.46 \pm 0.29$ & $90.79 \pm 0.17$ & $91.31 \pm 0.04$ \\
\bottomrule
\end{tabular}
\label{tab:CIFAR_classification_results}
\end{table}

The results in Table \ref{tab:NLP_classification_results} mirror the findings from the vision tasks. Both PARAM configurations outperform the baseline across all evaluated language pairs. For both modalities, the framework achieves zero-shot performance levels that are comparable to the absolute classification results obtained in the original latent spaces.

\begin{table}[ht!]
\caption{Stitching performance with different language-based encoders. The table shows the mean-weighted F1 ($\pm$ std) across $6$ seeds. Run with 300 anchors.}
\centering
\footnotesize
\setlength{\tabcolsep}{5pt}
\begin{tabular}{ll c ccc}
\toprule
& & \multicolumn{1}{c}{Absolute} & \multicolumn{3}{c}{Relative} \\
\cmidrule(lr){3-3}\cmidrule(lr){4-6}
Encoder & Classifier &
Abs\textbf{.} &
\makecell{Random-Cos} &
\multicolumn{2}{c}{PARAM} \\
\cmidrule(lr){5-6}
& & & & Multi & Single \\
\midrule

\multirow{5}{*}{fr} 
    & de & -- & $67.35 \pm 3.25$ & $92.17 \pm 0.32$ & $89.90 \pm 0.53$ \\
    & en & -- & $68.64 \pm 5.31$ & $92.02 \pm 0.23$ & $90.65 \pm 0.37$ \\
    & es & -- & $67.44 \pm 2.70$ & $92.06 \pm 0.34$ & $90.95 \pm 0.53$ \\
    & fr & $93.76 \pm 0.14$ & $92.06 \pm 0.26$ & $93.75 \pm 0.16$ & $93.59 \pm 0.40$ \\
    & zh & -- & $69.06 \pm 3.83$ & $92.26 \pm 0.15$ & $90.93 \pm 0.11$ \\
\midrule
\multirow{5}{*}{en} 
    & de & -- & $72.36 \pm 2.68$ & $87.89 \pm 0.38$ & $85.20 \pm 0.85$ \\
    & en & $90.25 \pm 0.36$ & $87.67 \pm 0.20$ & $89.70 \pm 0.10$ & $89.17 \pm 0.22$ \\
    & es & -- & $72.91 \pm 5.70$ & $87.82 \pm 0.25$ & $86.19 \pm 0.27$ \\
    & fr & -- & $75.88 \pm 4.05$ & $88.22 \pm 0.15$ & $84.43 \pm 0.74$ \\
    & zh & -- & $71.66 \pm 7.88$ & $88.28 \pm 0.28$ & $86.56 \pm 0.39$ \\
\bottomrule
\end{tabular}
\label{tab:NLP_classification_results}
\end{table}

The SLM nearest-neighbor retrieval evaluations seen in Table \ref{tab:SLM_stitching_results} reveal a distinct separation in performance between the proposed framework and the standard baseline. The Random-Cosine baseline demonstrates significant instability and a lack of alignment across the tested model pairings.

\begin{table}[t]
\caption{Stitching performance with small language models between a Source and Zero-shot encoder. Shows retrieval at one (R@1) and five (R@5) on the generated dataset across different seeds.}
\centering
\footnotesize
\setlength{\tabcolsep}{5pt}
\begin{tabular}{ll cccc}
\toprule
& & \multicolumn{2}{c}{Random--Cosine} & \multicolumn{2}{c}{PARAM--WIP} \\
\cmidrule(lr){3-4} \cmidrule(lr){5-6}
Source Enc& ZS Encoder& R@1 & R@5 & R@1 & R@5 \\
\midrule
\multirow{3}{*}{\texttt{TinyLlama}} & \texttt{gpt2} & $0.50 \pm 0.18$ & $2.28 \pm 0.49$ & $93.44 \pm 0.34$ & $99.94 \pm 0.14$ \\
 & \texttt{gpt2-med}& $0.39 \pm 0.14$ & $1.83 \pm 0.18$ & $94.28 \pm 0.53$ & $99.94 \pm 0.14$ \\
 & \texttt{gemma-3-1b} & $9.94 \pm 1.02$ & $28.33 \pm 3.15$ & $87.61 \pm 0.39$ & $98.67 \pm 0.63$ \\
\midrule
\multirow{3}{*}{\texttt{gpt2}} & \texttt{TinyLlama} & $1.72 \pm 0.25$ & $6.06 \pm 0.44$ & $95.67 \pm 0.42$ & $100.00 \pm 0.00$ \\
 & \texttt{gpt2-med}& $0.44 \pm 0.17$ & $2.17 \pm 0.66$ & $94.67 \pm 0.37$ & $100.00 \pm 0.00$ \\
 & \texttt{gemma-3-1b} & $1.83 \pm 0.28$ & $5.11 \pm 0.86$ & $87.61 \pm 0.57$ & $99.33 \pm 0.47$ \\
\bottomrule
\end{tabular}
\label{tab:SLM_stitching_results}
\end{table}

\section{Ablation Study}
\label{sec:ablation}
To determine whether anchor construction (Random vs.\ \PARAM) or the similarity measure (Cosine vs.\ \WIP) drives performance, we evaluate zero-shot accuracy alongside three representational metrics: Mean Reciprocal Rank (MRR), normalized Mean Squared Error (MSE), and the standard deviation of relative features ($RR_{std}$). Following the zero-shot protocol in Algorithm \ref{alg:zs_pipeline}, we utilize two source spaces for training and one held-out target space for evaluation. In our experiments, we find that a moderate support set is sufficient, using roughly 5--15 points per embedding dimension already reaches near-saturated zero-shot performance on CIFAR-100 (coarse) (Appendix~\ref{app:subset_size}).

\subsection{Ablation Results}

\begin{figure}[t]
  \centering
  \begin{subfigure}[t]{0.48\linewidth}
    \centering
    \includegraphics[width=\linewidth]{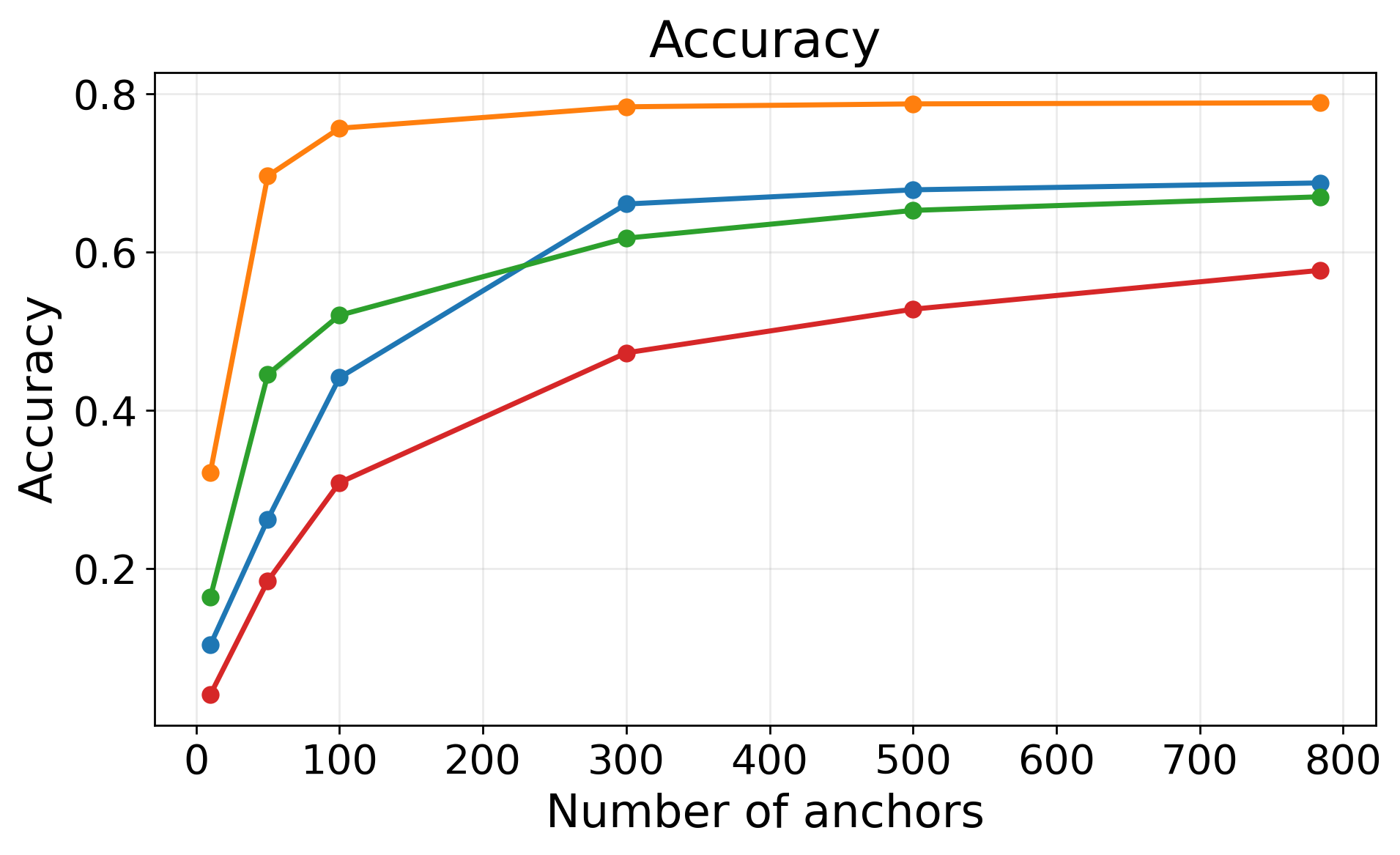}
    \caption{Accuracy}
    \label{fig:ablation_acc}
  \end{subfigure}\hfill
  \begin{subfigure}[t]{0.48\linewidth}
    \centering
    \includegraphics[width=\linewidth]{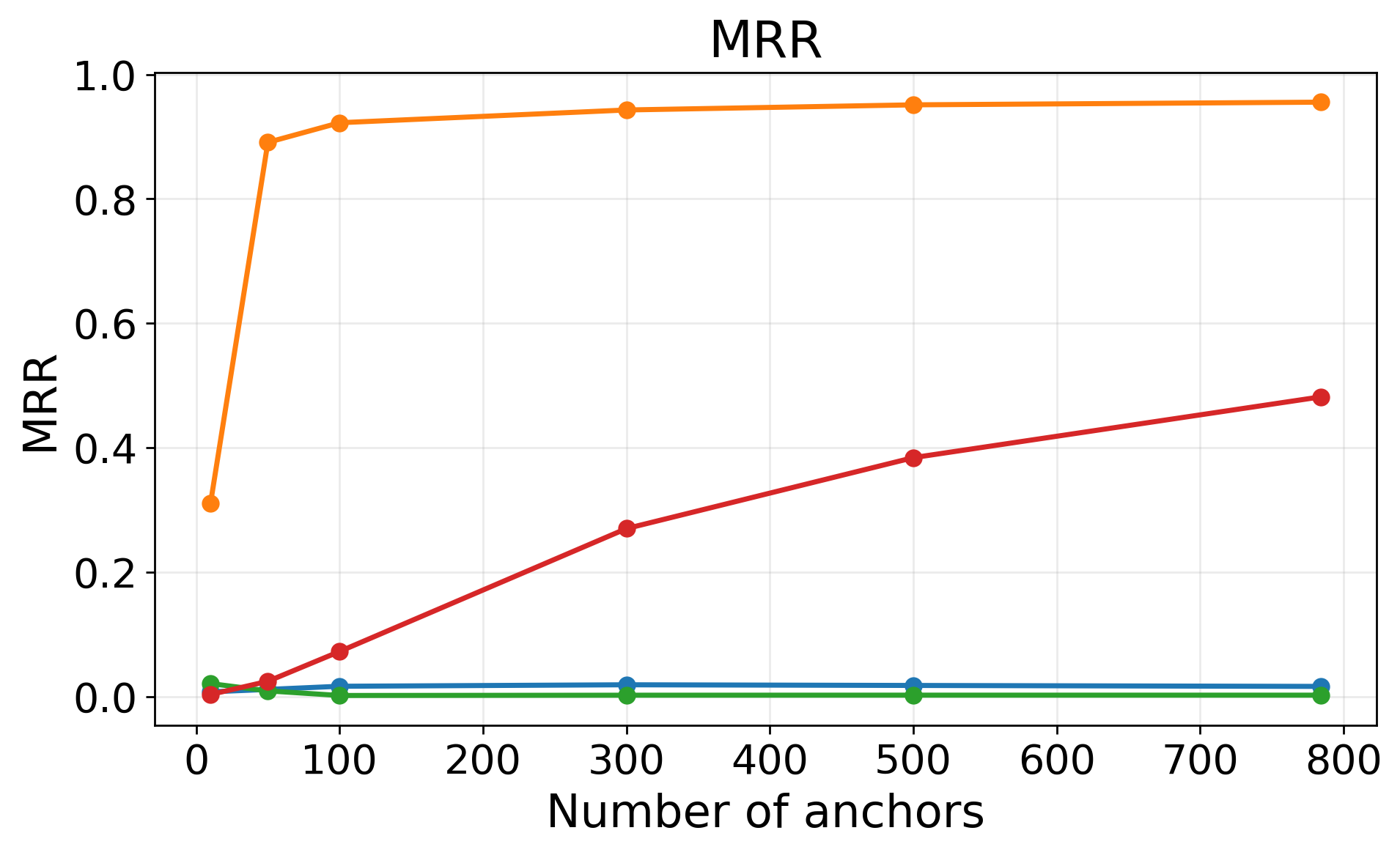}
    \caption{MRR}
    \label{fig:ablation_mrr}
  \end{subfigure}

  \vspace{-3pt}
  \begin{subfigure}[t]{\linewidth}
    \centering
    \includegraphics[width=0.7\linewidth]{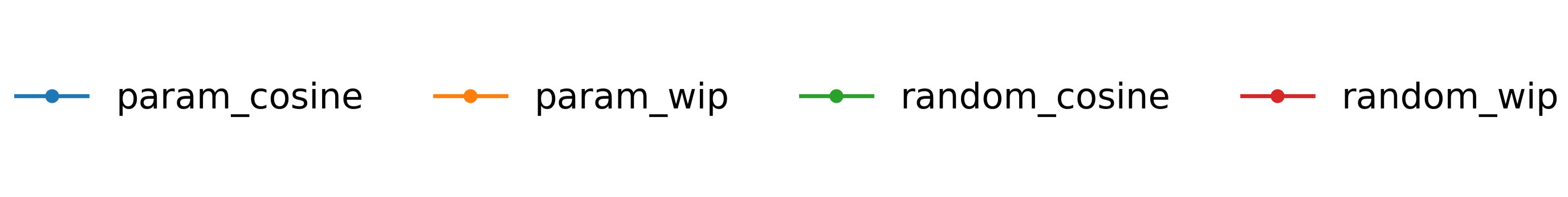}
    \label{fig:ablation_legend}
  \end{subfigure}
  \vspace{-20pt}
  \caption{Ablation across anchor counts for anchor construction (random vs.\ \PARAM) and similarity (cosine vs.\ \WIP).}
  \label{fig:ablation_plots}
\end{figure}

Figure \ref{fig:ablation_plots} shows that \PARAM-\WIP consistently achieves the best zero-shot accuracy across all anchor counts. While \WIP significantly improves retrieval quality (MRR), its largest accuracy gains appear only when combined with learned anchors. Conversely, cosine similarity performs similarly regardless of anchor construction, acting as the primary bottleneck in representational alignment.

Notably, the Random-WIP configuration often underperforms Random-Cosine. This may be due to random anchors having a higher probability of being sampled around the origin when the space is isometric. Since WIP maintains magnitude information, this may lead to several low-magnitude dimensions with high variance between them, which results in higher sensitivity to noise and worse alignment between spaces.

\subsection{PARAM Single vs Multi}
We further compare \PARAM trained with single-space losses to the multi-space variant utilizing $\mathcal{L}_{symNCE}$. As shown in Table \ref{tab:rr_metrics_comparison}, multi-space training substantially improves cross-space alignment by increasing MRR and reducing normalized MSE (MSE divided by the variance of the RR features). The metrics also show that \PARAM Multi and Single both maintain a much higher feature spread ($RR_{std}$) compared to the baseline.

\begin{table}[t]
\centering
\caption{RR metrics comparing \PARAM Single, \PARAM Multi, and Random+Cosine across \texttt{ViT-B} and \texttt{ViT-S}. Mean $\pm$ std over seeds.}
\label{tab:rr_metrics_comparison}
\small
\begin{tabular*}{\textwidth}{@{\extracolsep{\fill}}lccc@{}}
\toprule
Metric & \PARAM Single & \PARAM Multi & Random+Cosine \\ \midrule
MRR           & $0.281 \pm 0.003$   & $0.396 \pm 0.002$  & $0.016 \pm 0.001$    \\
MSE (normed)  & $0.961 \pm 0.006$   & $0.427 \pm 0.002$  & $1.108 \pm 0.020$    \\
$RR_{std}$    & $1.213 \pm 0.001$   & $1.103 \pm 0.003$  & $0.067 \pm 0.001$    \\ 
\bottomrule
\end{tabular*}
\end{table}

\section{Discussion}
The performance gap between our framework and the standard baseline reveals a fundamental limitation in current relative representations: the reliance on random sampling and angular projection. Our results indicate that the Random-Cosine baseline is inherently unsuitable for heterogeneous model families, particularly in language model stitching, where it exhibits extreme instability. This failure suggests that cosine similarity acts as a lossy filter that is overly sensitive to anisotropic latent spaces, e.g. those of transformers. In contrast, WIP manages these geometric discrepancies better by providing an affine-invariant function that preserves both magnitude and angular information, satisfying the objective of Cross-Space Alignment ($O_{al}$) without discarding discriminative information ($O_{inf}$).

Furthermore, the stability of PARAM-based relative spaces suggests that learned anchors prevent the dimensionality collapse inherent in random sampling. By optimizing across multiple encoders, we produce anchors that act as universal semantic prototypes, ensuring that the relative mapping, $RR(x)$, retains task-relevant information in the absolute embeddings. The nearly lossless transfer observed in our classification tasks -- where zero-shot performance matches absolute performance -- confirms that PARAM-WIP enables a robust, universal relative space that is remarkably consistent across independently trained encoders.

\subsection{Limitations and future work}
\label{sec:discussion}
\emph{Limitations:} 
(1) \PARAM anchors lie in the convex hull of the subset embeddings; representational capacity depends on subset size and diversity.
(2) Whitening requires stable covariance estimation; shrinkage helps but adds hyperparameters ($\lambda$, $\varepsilon$).
(3) RR stitching is not guaranteed under severe nonlinear distortions between spaces; our approach targets common linear/affine mismatches and anisotropy.

\emph{Future work:} 
Extensions include adaptive subset selection, structured sparsity on $\mathbf{P}$ for interpretability and speed, and exploring more paired and unpaired alignment objectives at scale, as well as extending RR with \emph{non-linear} similarity functions or learnable warps (e.g., shallow MLPs, kernelized similarities,
or normalizing-flow style mappings) while preserving the “shared-anchor” interface, potentially enabling stitching across more heterogeneous encoders.

\section{Conclusion}
\label{sec:conclusion}
We presented PARAM, a learnable anchor mechanism, and WIP, a geometry-aware similarity metric, designed to provide a reliable foundation for zero-shot latent space communication. By moving beyond static, randomly sampled anchors and affine-sensitive measures like cosine similarity, our framework addresses the geometric and informational bottlenecks of standard Relative Representations. Our results across disparate vision and language encoders, including large-scale language models, demonstrate significant gains in both performance and consistency. Ultimately, by formalizing the relationship between latent geometry and relative coordinate systems, we provide a practical and scalable pathway toward modular AI architectures where encoders and decoders can be interchanged without costly retraining or adaptation.

\bibliographystyle{unsrt}
\bibliography{main}
\newpage

\section*{Appendix}

\appendix
\section{PARAM Loss Definitions}
\label{app:losses}

\subsubsection{Single-Space Objectives} 
These terms are applied per training encoder $E_i$ and then averaged across encoders.

\paragraph{Coverage (soft k-means in whitened space).}
Let $x_w^i = L^i(x - \mu^i)$ and $a_{w,r}^i = L^i(a_r^i - \mu^i)$ denote the whitened embeddings and whitened anchors respectively for space $i$, where $L^i$ is the shrunk covariance estimate defined in Appendix \ref{app:wip}. For each whitened embedding $x_w^i$, we form a soft assignment to the $r$-th anchor:
\[
q_r^i(x) = \frac{\exp(-\|x_w^i - a_{w,r}^i\|_2^2 / \tau_{cov})}{\sum_{j=1}^{m} \exp(-\|x_w^i - a_{w,j}^i\|_2^2 / \tau_{cov})}.
\]
The coverage objective is the expected soft k-means distortion across the $d$ dimensions:
\[
\mathcal{L}_{\mathrm{cov}} = \mathbb{E}_{x} \Big[ \sum_{r=1}^{m} q_r^i(x) \|x_w^i - a_{w,r}^i\|_2^2 \Big] / d.
\]

\paragraph{Orthogonality (Effective Rank).} 
To promote diversity and prevent anchor collapse, we regularize the unit-normalized whitened anchors $\hat{a}_{w,r}^i = a_{w,r}^i / \|a_{w,r}^i\|_2$. Let $G^i \in \mathbb{R}^{m \times m}$ be their Gram matrix, where $r, s \in \{1, \dots, m\}$ denote the indices of anchors in the set:
\[
G_{rs}^i = (\hat{a}_{w,r}^i)^\top \hat{a}_{w,s}^i = \cos \angle(\hat{a}_{w,r}^i, \hat{a}_{w,s}^i).
\]
The orthogonality loss penalizes the mean squared cosine similarity over all distinct anchor pairs ($r \neq s$):
\[
\mathcal{L}_{\mathrm{orth}} = \frac{1}{m(m-1)} \sum_{r \neq s} (G_{rs}^i)^2 = \frac{\|G^i\|_{F}^{2} - m}{m(m-1)},
\]
where the minimum $\mathcal{L}_{\mathrm{orth}} = 0$ is attained when all anchors in the whitened space are mutually orthogonal.

\paragraph{Anchor length constraint.}
We constrain the anchors to maintain a controlled magnitude after whitening to prevent drift toward uninformative scales:
\[
\mathcal{L}_{\mathrm{len}} = \frac{1}{m} \sum_{r=1}^{m} \big( \|a_{w,r}^i\|_2 - 1 \big)^2.
\]

\subsubsection{Multi-Space Objectives}
Let $R^i \in \mathbb{R}^{B \times m}$ and $R^j \in \mathbb{R}^{B \times m}$ be minibatches of Relative Representations (RR) from encoders $E_i$ and $E_j$, where rows correspond to parallel samples $x_1, \dots, x_B$.

\paragraph{Symmetric InfoNCE (Information Noise-Contrastive Estimation).}
We define the cosine logits between the $a$-th sample from space $i$ and the $b$-th sample from space $j$:
\[
\ell_{ab} = \frac{\langle \hat{r}^i(x_a), \hat{r}^j(x_b) \rangle}{\tau_{NCE}}, \qquad \hat{r} = \frac{RR(x)}{\|RR(x)\|_2}.
\]
The one-positive cross-entropy loss for the direction $i \to j$ is given by:
\[
\mathcal{L}_{\mathrm{NCE}}(i \to j) = -\frac{1}{B} \sum_{a=1}^{B} \log \frac{\exp(\ell_{aa})}{\sum_{b=1}^{B} \exp(\ell_{ab})}.
\]
The final alignment is enforced via the symmetric variant:
\[
\mathcal{L}_{\mathrm{symNCE}}(i, j) = \frac{1}{2} \Big( \mathcal{L}_{\mathrm{NCE}}(i \to j) + \mathcal{L}_{\mathrm{NCE}}(j \to i) \Big).
\]

\section{Whitened Inner Product (WIP): Definition and Affine Invariance}
\label{app:wip}

For embedding space $i$, let $\mu^i$ and $\hat{\Sigma}^i$ be the empirical mean and covariance estimated from a metric set (typically the full embedding set for stability if available). We use the shrinkage-stabilized covariance matrix $\Sigma_\lambda^i$ as defined in the main text:
\[
\Sigma_\lambda^i = (1-\lambda)\hat{\Sigma}^i + \lambda\frac{\mathrm{tr}(\hat{\Sigma}^i)}{d}I + \epsilon I,
\qquad
L^i = (\Sigma_\lambda^i)^{-1/2},
\]
where $\lambda \in [0,1]$ is a linear shrinkage coefficient that interpolates between the sample covariance and an isotropic target $\frac{\mathrm{tr}(\hat{\Sigma}^i)}{d}I$, $d$ is the embedding dimension, and $\epsilon > 0$ is a small constant for numerical stability. In all experiments, we set $\lambda = 0.15$ and $\epsilon = 5 \times 10^{-8}$. For an embedding $x$ and anchor $a$ in space $i$, the WIP similarity is:
\[
s_{\mathrm{WIP}}(x,a) = \langle L^i(x-\mu^i), L^i(a-\mu^i) \rangle = (x-\mu^i)^\top(\Sigma_\lambda^i)^{-1}(a-\mu^i).
\]

\subsection*{Affine invariance (non-shrunk case)}

Consider an affine transform $y = Ax + c$ with invertible $A \in \mathbb{R}^{d \times d}$ and $c \in \mathbb{R}^d$ (this includes rotations and anisotropic scalings). Let $\mu^y = A\mu^x + c$ and $\Sigma^y = A\Sigma^x A^\top$. For $\lambda=0$ and $\epsilon=0$, the similarity remains invariant:
\[
(y-\mu^y)^\top (\Sigma^y)^{-1} (b-\mu^y) = (x-\mu^x)^\top (\Sigma^x)^{-1} (a-\mu^x),
\]
where $b = Aa + c$ is the transformed anchor. Proof:
\begin{align*}
    (y-\mu^y)^\top (\Sigma^y)^{-1} (b-\mu^y) = (A(x-\mu^x))^\top (A\Sigma^x A^\top)^{-1} (A(a-\mu^x)) \\
    = (x-\mu^x)^\top (\Sigma^x)^{-1} (a-\mu^x)
\end{align*}

Thus, the non-shrunk WIP score is invariant to any invertible affine transform. In practice, we use shrinkage ($\lambda > 0$) and the additive $\epsilon I$ for numerical stability; this can slightly break exact invariance under general $A$ but improves robustness in high-dimensional settings.

\section{Support-set size and diminishing returns}
\label{app:subset_size}

\begin{figure}[h!]
  \centering
  \includegraphics[width=\linewidth]{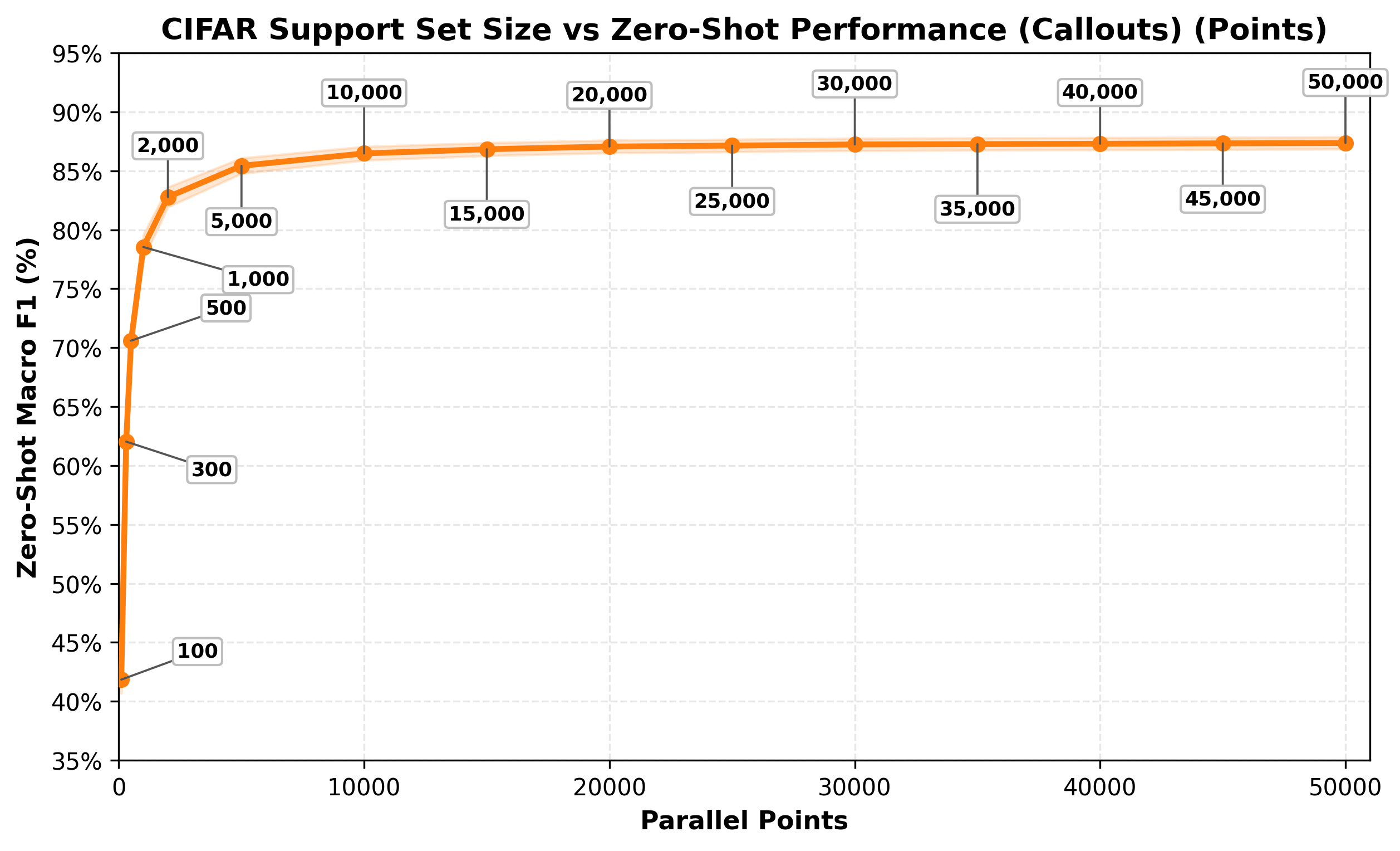}
  
  \caption{Effect of support-set size ($|\mathbf{X}_{\text{sub}}|$) on CIFAR-100 zero-shot performance. After roughly $5{,}000$--$10{,}000$ parallel points ($\approx$10--20\% of the dataset), gains diminish substantially, with performance saturating as more points are added.}
  \label{fig:subset_size_vs_f1}
\end{figure}

Fig.~\ref{fig:subset_size_vs_f1} shows how zero-shot performance depends on the number of parallel points used to construct the shared support set $X_{\text{sub}}$.
With very small support sets, anchor construction is underconstrained and stitching quality degrades.
However, performance improves rapidly as the support set grows and then saturates: beyond approximately $5{,}000$--$10{,}000$ points ($\approx$10--20\% of the CIFAR-100 train set size), adding more parallel points yields only marginal gains.
In practice, this suggests that \PARAM can achieve near-peak stitching performance using a moderate support set, reducing both memory and compute without substantially affecting results.

\section{Implementation Details and Hyperparameters}
\label{app:hyperparams}

\paragraph{Implementation details.}
We implement the \PARAM mixture matrix $\mathbf{P}$ as unconstrained logits and apply a row-wise softmax to enforce a row-stochastic convex mixture.
For numerical stability in high-dimensional settings, whitening matrices are computed via a stabilized eigendecomposition in \texttt{float64} precision before being cast to the model's working precision.
All RR computations and soft-assignment terms are computed in chunks/blocks to limit memory usage.

\paragraph{Hyperparameters.}
Table~\ref{tab:hyperparams} lists the hyperparameters used in our main experiments. The hyperparameters were selected based on alignment metrics (e.g., neighborhood overlap and effective rank) on a representative validation subset of encoders, ensuring the learned anchors generalize across latent geometries.

\begin{table}[h]
\caption{Hyperparameters used for \PARAM--\WIP training and evaluation.}
\centering
\setlength{\tabcolsep}{10pt}
\begin{tabular}{@{}ll@{}}
\toprule
Hyperparameter & Value \\
\midrule
Shrinkage ($\lambda$) & $0.15$ \\
Covariance epsilon ($\epsilon$) & $5.0\times10^{-8}$ \\
Softmax temperature for $P$ ($T$) & $2.9$ \\
Epochs & $95$ \\
Learning rate (Adam/AdamW) & $0.03$ \\
Batch size & $1024$ \\
Coverage chunk size & $16384$ \\
Contrastive weight ($w_{\mathrm{symNCE}}$) & $0.70$ \\
Contrastive temperature ($\tau$) & $0.10$ \\
Coverage weight ($w_{\mathrm{cov}}$) & $11.3$ \\
Coverage temperature ($\tau_{\mathrm{cov}}$) & $2.7$ \\
Anchor orthogonality weight ($w_{\mathrm{orth}}$) & $0.27$ \\
Anchor length weight ($w_{\mathrm{len}}$) & $1.2$ \\
\bottomrule
\end{tabular}
\label{tab:hyperparams}
\end{table}

\section{Additional Results}

\subsection{Full Zero-Shot Text Classification Results}
Table \ref{tab:NLP_classification_results_full} shows the full results table for the amazon classification task.

\begin{table}[ht!!]
\caption{Extension of the results from Section \ref{sec:results}. The table shows the mean-weighted F1 ($\pm$ std) across $6$ seeds. Run with 300 anchors.}
\centering
\footnotesize
\setlength{\tabcolsep}{5pt}
\begin{tabular}{ll c ccc}
\toprule
& & \multicolumn{1}{c}{Absolute} & \multicolumn{3}{c}{Relative} \\
\cmidrule(lr){3-3}\cmidrule(lr){4-6}
Encoder & Classifier & 
Abs. & 
\makecell{Random-Cos} & 
\multicolumn{2}{c}{PARAM} \\
\cmidrule(lr){5-6}
& & & & Multi & Single \\
\midrule

\multirow{5}{*}{de} 
    & de & $89.20 \pm 0.16$ & $85.92 \pm 0.26$ & $88.12 \pm 0.17$ & $87.69 \pm 0.30$ \\
    & en & --               & $74.18 \pm 2.54$ & $86.40 \pm 0.35$ & $84.56 \pm 0.43$ \\
    & es & --               & $75.09 \pm 3.80$ & $85.91 \pm 0.35$ & $84.44 \pm 0.33$ \\
    & fr & --               & $73.79 \pm 2.29$ & $86.24 \pm 0.22$ & $82.64 \pm 0.46$ \\
    & zh & --               & $70.39 \pm 4.10$ & $86.60 \pm 0.20$ & $84.79 \pm 0.18$ \\
\midrule
\multirow{5}{*}{en} 
    & de & --               & $72.36 \pm 2.68$ & $87.89 \pm 0.38$ & $85.20 \pm 0.85$ \\
    & en & $90.25 \pm 0.36$ & $87.67 \pm 0.20$ & $89.70 \pm 0.10$ & $89.17 \pm 0.22$ \\
    & es & --               & $72.91 \pm 5.70$ & $87.82 \pm 0.25$ & $86.19 \pm 0.27$ \\
    & fr & --               & $75.88 \pm 4.05$ & $88.22 \pm 0.15$ & $84.43 \pm 0.74$ \\
    & zh & --               & $71.66 \pm 7.88$ & $88.28 \pm 0.28$ & $86.56 \pm 0.39$ \\
\midrule
\multirow{5}{*}{es} 
    & de & --               & $76.83 \pm 1.77$ & $88.89 \pm 0.23$ & $86.94 \pm 0.84$ \\
    & en & --               & $75.21 \pm 3.13$ & $89.16 \pm 0.28$ & $87.81 \pm 0.30$ \\
    & es & $91.89 \pm 0.25$ & $89.18 \pm 0.24$ & $91.36 \pm 0.04$ & $90.85 \pm 0.22$ \\
    & fr & --               & $77.79 \pm 3.74$ & $89.39 \pm 0.29$ & $86.18 \pm 0.50$ \\
    & zh & --               & $77.00 \pm 3.18$ & $89.71 \pm 0.25$ & $87.92 \pm 0.33$ \\
\midrule
\multirow{5}{*}{fr} 
    & de & --               & $67.35 \pm 3.25$ & $92.17 \pm 0.32$ & $89.90 \pm 0.53$ \\
    & en & --               & $68.64 \pm 5.31$ & $92.02 \pm 0.23$ & $90.65 \pm 0.37$ \\
    & es & --               & $67.44 \pm 2.70$ & $92.06 \pm 0.34$ & $90.95 \pm 0.53$ \\
    & fr & $93.76 \pm 0.14$ & $92.06 \pm 0.26$ & $93.75 \pm 0.16$ & $93.59 \pm 0.40$ \\
    & zh & --               & $69.06 \pm 3.83$ & $92.26 \pm 0.15$ & $90.93 \pm 0.11$ \\
\midrule
\multirow{5}{*}{zh} 
    & de & --               & $78.91 \pm 2.41$ & $86.42 \pm 0.34$ & $83.88 \pm 0.56$ \\
    & en & --               & $81.21 \pm 1.49$ & $86.88 \pm 0.21$ & $84.53 \pm 0.47$ \\
    & es & --               & $79.70 \pm 2.31$ & $86.21 \pm 0.23$ & $84.95 \pm 0.66$ \\
    & fr & --               & $81.17 \pm 1.78$ & $86.58 \pm 0.15$ & $83.98 \pm 0.45$ \\
    & zh & $88.42 \pm 0.12$ & $87.13 \pm 0.17$ & $88.04 \pm 0.17$ & $87.18 \pm 0.26$ \\
\bottomrule
\end{tabular}
\label{tab:NLP_classification_results_full}
\end{table}

\subsection{Full Zero-Shot Text Classification Results}
Table \ref{tab:SLM_stitching_results_full} shows the full results table for the SLM retrieval task.

\makeatletter
\setlength{\@fptop}{0pt}
\setlength{\@fpsep}{8pt}
\setlength{\@fpbot}{0pt plus 1fil}
\makeatother

\begin{table}[!t]
\caption{Extension of the results from Section \ref{sec:results}. Shows stitching performance with small language models. Mean retrieval accuracy (\%) ($\pm$ std) across seeds. Shows retrieval at one (R@1) and five (R@5) accuracy on the generated dataset.}
\centering
\footnotesize
\setlength{\tabcolsep}{5pt}
\begin{tabular}{ll cccc}
\toprule
& & \multicolumn{2}{c}{Random--Cosine} & \multicolumn{2}{c}{PARAM--WIP} \\
\cmidrule(lr){3-4} \cmidrule(lr){5-6}
Source Enc& ZS Encoder& R@1 & R@5 & R@1 & R@5 \\
\midrule
\multirow{3}{*}{\texttt{TinyLlama}} & \texttt{gpt2} & $0.50 \pm 0.18$ & $2.28 \pm 0.49$ & $93.44 \pm 0.34$ & $99.94 \pm 0.14$ \\
 & \texttt{gpt2-med} & $0.39 \pm 0.14$ & $1.83 \pm 0.18$ & $94.28 \pm 0.53$ & $99.94 \pm 0.14$ \\
 & \texttt{gemma-3-1b} & $9.94 \pm 1.02$ & $28.33 \pm 3.15$ & $87.61 \pm 0.39$ & $98.67 \pm 0.63$ \\
\midrule
\multirow{3}{*}{\texttt{gpt2}} & \texttt{TinyLlama} & $1.72 \pm 0.25$ & $6.06 \pm 0.44$ & $95.67 \pm 0.42$ & $100.00 \pm 0.00$ \\
 & \texttt{gpt2-med}& $0.44 \pm 0.17$ & $2.17 \pm 0.66$ & $94.67 \pm 0.37$ & $100.00 \pm 0.00$ \\
 & \texttt{gemma-3-1b} & $1.83 \pm 0.28$ & $5.11 \pm 0.86$ & $87.61 \pm 0.57$ & $99.33 \pm 0.47$ \\
\midrule
\multirow{3}{*}{\texttt{gpt2-med}}& \texttt{TinyLlama} & $1.17 \pm 0.18$ & $5.11 \pm 0.27$ & $95.72 \pm 0.25$ & $100.00 \pm 0.00$ \\
 & \texttt{gpt2} & $0.78 \pm 0.17$ & $4.22 \pm 0.34$ & $94.94 \pm 0.25$ & $100.00 \pm 0.00$ \\
 & \texttt{gemma-3-1b} & $1.61 \pm 0.65$ & $6.83 \pm 1.38$ & $88.06 \pm 0.25$ & $99.22 \pm 0.27$ \\
\midrule
\multirow{3}{*}{\texttt{gemma-3-1b}} & \texttt{TinyLlama} & $0.94 \pm 0.33$ & $3.44 \pm 0.58$ & $84.22 \pm 1.52$ & $97.72 \pm 0.33$ \\
 & \texttt{gpt2} & $0.33 \pm 0.00$ & $1.61 \pm 0.14$ & $81.61 \pm 0.65$ & $97.44 \pm 0.54$ \\
 & \texttt{gpt2-med}& $0.33 \pm 0.00$ & $1.67 \pm 0.00$ & $83.39 \pm 1.71$ & $97.44 \pm 0.40$ \\
\bottomrule
\end{tabular}
\label{tab:SLM_stitching_results_full}
\end{table}

\clearpage

\subsection{Image Reconstruction with Anisotropic Spaces}

\paragraph{Experimental setting.} To evaluate the robustness of relative mappings under significant geometric distortions, we perform an MNIST reconstruction task, training four independent autoencoders ($AE_1, \dots, AE_4$) using distinct seeds. To simulate a \emph{challenging zero-shot scenario}, we intentionally induce anisotropic scaling during the training, resulting in highly dissimilar latent geometries. A relative decoder is trained on embeddings from $AE_1$ and is stitched with the RR of embeddings from $AE_4$. We evaluate the performance of PARAM Multi against the Random-Cosine baseline. This setup tests the hypothesis that a robust relative interface should facilitate successful zero-shot transfer if task-relevant semantic information is preserved in the absolute latent spaces.

\begin{figure}[!t]
    \centering
    \includegraphics[width=1\linewidth]{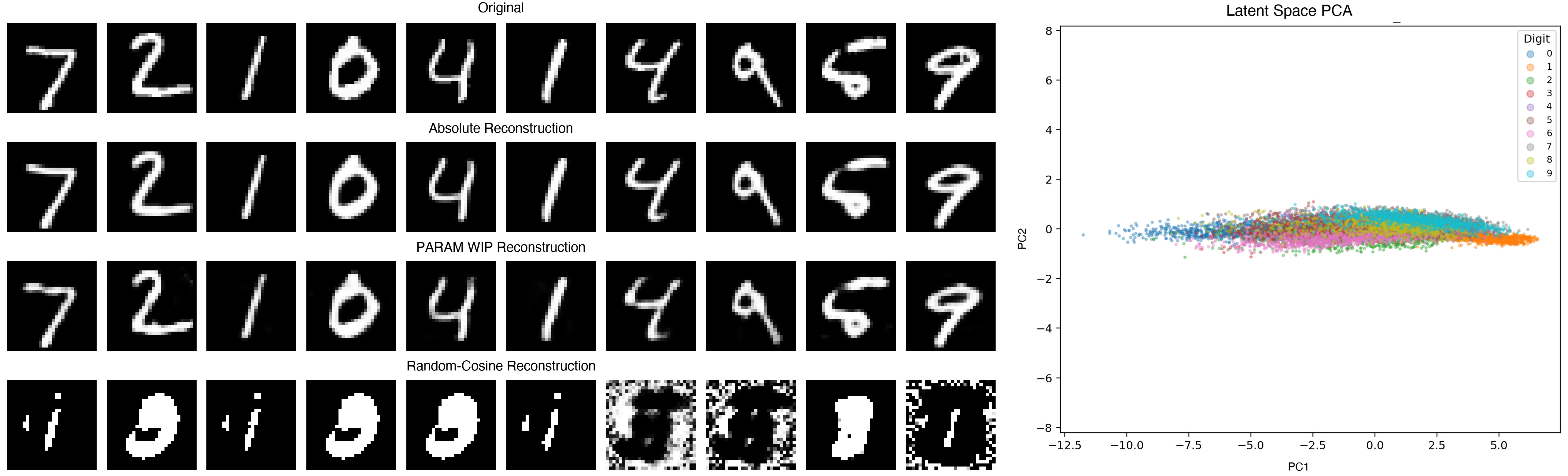}
    \caption{MNIST reconstruction with an anisotropic latent space. The bottom 2 rows are zero-shot stitching using RR. The PCA plot (right) illustrates the induced "cigar-shaped" latent distribution. Latent space dim and number of anchors are set to 64.}
    \label{fig:MNIST_recon}
\end{figure}

\paragraph{Results analysis.} Figure~\ref{fig:MNIST_recon} demonstrates a significant performance gap in this induced stress test under extreme latent anisotropy. While the absolute reconstruction remains the reference, the Random-Cosine baseline fails almost entirely, producing noisy and unrecognizable outputs. This failure is caused by the induced anisotropic scaling shown in the PCA plot; cosine similarity's normalization discards essential magnitude and directional information required for "cigar-shaped" distributions. In contrast, PARAM WIP generates high-fidelity reconstructions nearly identical to the absolute baseline.

\end{document}